\documentclass{article}

\usepackage{arxiv}

\usepackage[utf8]{inputenc} 
\usepackage[T1]{fontenc}    
\usepackage{hyperref}       
\usepackage{url}            
\usepackage{booktabs}       
\usepackage{amsfonts}       
\usepackage{nicefrac}       
\usepackage{microtype}      
\usepackage{lipsum}
\usepackage{graphicx}
\usepackage{amsmath,amssymb,amsfonts}
\usepackage{algorithmic}
\usepackage{graphicx}
\usepackage{textcomp}
\usepackage{booktabs}
\usepackage{multirow}
\usepackage{hyperref}
\graphicspath{ {./images/} }
\usepackage{natbib}

\newcommand{\mycomment}[1]{}

\title{UniBERT: Adversarial Training for Language-Universal Representations}

\author{Andrei-Marius Avram\textsuperscript{1,3},
Marian Lupa\c{s}cu\textsuperscript{2,3}, Dumitru-Clementin Cercel\textsuperscript{1}\thanks{Corresponding author: dumitru.cercel@upb.ro.}, \\ \textbf{Ionu\c{t} Mironic\u{a}\textsuperscript{3}},  \textbf{\c{S}tefan Tr\u{a}usan-Matu\textsuperscript{1,4}}\\
\textsuperscript{1}National University of Science and Technology POLITEHNICA Bucharest, Bucharest, Romania \\
\textsuperscript{2}University of Bucharest, Bucharest, Romania \\
\textsuperscript{3}Adobe Research, Bucharest, Romania \\
\textsuperscript{4}Academy of Romanian Scientists (AOSR), Bucharest, Romania \\
}

\begin{document}
\maketitle
\begin{abstract}
This paper presents UniBERT, a compact multilingual language model that uses an innovative training framework that integrates three components: masked language modeling, adversarial training, and knowledge distillation. Pre-trained on a meticulously curated Wikipedia corpus spanning 107 languages, UniBERT is designed to reduce the computational demands of large-scale models while maintaining competitive performance across various natural language processing tasks. Comprehensive evaluations on four tasks—named entity recognition, natural language inference, question answering, and semantic textual similarity—demonstrate that our multilingual training strategy enhanced by an adversarial objective significantly improves cross-lingual generalization. Specifically, UniBERT models show an average relative improvement of 7.72\% over traditional baselines, which achieved an average relative improvement of only 1.17\%, and statistical analysis confirms the significance of these gains (p-value = 0.0181). This work highlights the benefits of combining adversarial training and knowledge distillation to build scalable and robust language models, thus advancing the field of multilingual and cross-lingual natural language processing.
\end{abstract}

\section{Introduction}
\label{sec:introduction}
Natural language processing (NLP) has advanced in recent years due to breakthroughs in deep learning and Transformer-based architectures \citep{vaswani2017attention}. Models such as Bidirectional Encoder Representations from Transformers (BERT) \citep{devlin-etal-2019-bert} and its successors \citep{liu2019roberta,clark2020electra} have redefined the state of the art in tasks such as machine translation, sentiment analysis, summarization, and question answering by learning contextualized representations that capture both syntactic and semantic nuances. This evolution has not only elevated task-specific performance, but also opened new avenues for exploring the complexities of human language.

At the same time, there has been a surge of interest in developing multilingual systems capable of processing and understanding multiple languages simultaneously \cite{conneau-etal-2020-unsupervised,xue-etal-2021-mt5,yano2024multilingual}. Multilingual training leverages shared linguistic features across diverse languages, a benefit that is especially pronounced in low-resource settings. In addition, adversarial training \cite{43405} has emerged as a powerful mechanism to encourage the learning of language-invariant representations, thus reducing language-specific biases and improving cross-lingual generalization in a variety of tasks \cite{avram2023multilingual,avram2025rolargesum}.

However, developing robust multilingual models poses several challenges, including managing heterogeneous data distributions, accommodating language-specific nuances, and ensuring scalability without compromising performance. Adversarial training addresses these difficulties by introducing a competitive dynamic that forces the model to focus on features common to all languages. This approach not only boosts overall performance but also diminishes the model’s reliance on language-specific signals to enable better generalization across many linguistic domains \cite{xu2024survey}.

In this work, we introduce UniBERT, a compact and efficient BERT-based model specifically designed for multilingual processing,  available in three open-source versions: (1) UniBERT-Small\footnote{https://huggingface.co/avramandrei/unibert-small.}, (2) UniBERT-XSmall\footnote{https://huggingface.co/avramandrei/unibert-xsmall.}, and (3) UniBERT-XXSmall\footnote{https://huggingface.co/avramandrei/unibert-xxsmall.}. UniBERT was pre-trained on a curated Wikipedia corpus spanning 107 languages and combines three key training components: masked language modeling \cite{devlin-etal-2019-bert}, adversarial training, and knowledge distillation \cite{hinton2015distilling}. This integrated methodology enables the model to achieve robust performance while significantly reducing the parameter count relative to larger models, making it an attractive solution for resource-constrained applications.

We perform extensive evaluations of UniBERT on four NLP tasks: (1) named entity recognition, (2) natural language inference, (3) question answering, and (4) semantic textual similarity. The results demonstrate that our models, fine-tuned jointly on all languages, obtain better results than when they are fine-tuned on each language individually. Concretely, on average, UniBERT-Small improves the performance from 47.10\% to 48.87\%, UniBERT-XSmall from 40.20\% to 44.35\%, and UniBERT-XXSmall from 34.51\% to 37.64\% when trained jointly in all languages, yielding an average relative improvement of 7.72\%. In contrast, larger language models show an improvement of 1.12\%, underscoring the significant benefits of the multilingual joint training approach enhanced by adversarial objectives. 

We also performed a comprehensive statistical comparative analysis to confirm the strengths of our approach. The student's t-test yielded a p-value of 0.0184, demonstrating that gains in UniBERT performance are significant at a significance threshold of $\alpha = 0.05$. It also shows that the combination of multilingual joint  training and adversarial training effectively compensates for the reduced capacity of compact models. Finally, we demonstrate the domain adaptability of UniBERT models through a comprehensive evaluation of legal documents, showing that our adversarial training approach facilitates effective knowledge transfer to specialized domains. In particular, the evaluation of the Multilingual Anonymisation for Public Administrations (MAPA) dataset \cite{DeGibertBonet2022} reveals remarkable improvements with relative gains of up to 55.71\% when using multilingual joint  training, further validating the effectiveness of our approach for cross-domain applications.

In summary, the main contributions of this paper are as follows:
\begin{itemize}
    \item We propose UniBERT, a compact, novel multilingual language model that integrates masked language modeling, adversarial training, and knowledge distillation.
    \item We show that multilingual joint  training, combined with an adversarial objective, significantly enhances cross-lingual performance, particularly in resource-constrained settings.
    \item We provide extensive evaluations on multiple NLP tasks, establishing the competitive performance of UniBERT compared to larger baseline models.
    \item We offer a detailed comparative analysis supported by statistical evidence, highlighting the efficacy of our approach in achieving language-invariant representations.
    \item We demonstrate the domain adaptability of UniBERT models through a comprehensive evaluation of legal documents, showing that our adversarial training approach facilitates effective knowledge transfer to specialized domains.
\end{itemize}

\section{Related Work}

\subsection{Multilingual Language Models}

The development of multilingual sentence embeddings has evolved from initial methods such as LASER \cite{chaudhary2019low}, which leveraged a shared long short-term memory \cite{gers2000learning} encoder-decoder framework to align sentence representations across languages using parallel data. Subsequent models, such as SBERT-distill \cite{reimers2020making}, utilized knowledge distillation to transfer monolingual sentence embedding capabilities into a multilingual space, allowing robust performance with fewer parallel sentences. LaBSE \cite{artetxe2019massively} expanded these ideas by introducing additive margin softmax, which improved the discriminative power for translation pair similarity. Despite their effectiveness, these methods often require extensive computational resources and large-scale parallel data, making them less accessible for low-resource languages.

Multilingual pre-trained language models have also made significant strides in cross-lingual understanding. Cross-lingual language
models (XLM) \cite{conneau2019cross} introduced the objective of translation language modeling (TLM), extending masked language modeling to leverage parallel data and align representations between languages. Building on this, XLM-RoBERTa (XLM-R) \cite{conneau-etal-2020-unsupervised} scaled these methods to 100 languages by pretraining on more than two terabytes of Common Crawl data \cite{wenzek-etal-2020-ccnet}, significantly improving performance on benchmarks for low-resource languages. XLM-R's success underscored the importance of large-scale data and careful language sampling strategies, particularly for balancing performance across high- and low-resource languages.

 mT5 \cite{xue-etal-2021-mt5}, which extended the T5 framework \cite{raffel2020exploring} to multilingual contexts, introduced further innovations. By adopting a text-to-text paradigm and pre-training using a large multilingual dataset covering 101 languages, mT5 achieved state-of-the-art results on various multilingual tasks. Balanced training was ensured by methods such as language sampling with controlled scaling and the use of large vocabularies. In conjunction with a span-corruption pretraining objective, the model's encoder-decoder design enabled flexibility across tasks ranging from classification to sequence generation.

Recent efforts, including EMS \cite{mao2024ems} and mSimCSE \cite{wang2022english}, have emphasized efficiency by reducing the reliance on large-scale parallel datasets and employing streamlined architectures. EMS's integration of token-level generative and sentence-level contrastive objectives exemplifies this shift, achieving a well-aligned multilingual embedding space with minimal computational overhead. Similarly, unsupervised approaches such as those used in SimCSE \cite{gao2021simcse} have been adapted for multilingual scenarios, providing competitive alternatives that further diminish the dependence on supervised data. In addition, further advances in the understanding of spoken language have emerged through multilingual mixture attention frameworks with adversarial training that adaptively capture unambiguous representations from aligned multilingual words while improving the scalability of the model for intent detection and slot filling tasks \cite{zhang2024multilingual}. These methods collectively push the boundaries of multilingual NLP, addressing challenges such as data scarcity and computational resource constraints.

Despite these advances, it is still difficult to scale these models to include underrepresented languages and optimize their performance in low-resource settings. Research on strategies such as improved sampling distributions, multilingual adversarial training, better representation of underrepresented data, and the integration of knowledge distillation continues to enhance the accessibility and efficacy of multilingual models \cite{xu2024survey}.

\subsection{Adversarial Training}

Adversarial training has become a cornerstone technique in machine learning, especially for improving model robustness and enhancing generalization to unseen data distributions. Initially introduced in the context of adversarial examples for image classification \cite{43405}, adversarial training has since been adapted to a variety of tasks, such as named entity recognition \cite{cai2024atbbc}, text classification \citep{lina2024adversarial,zhou2022rule}, and multiword expression detection \cite{avram2023romanian} in NLP. A central idea in adversarial training is the minimax optimization framework, which pits a discriminator against a feature extractor to encourage learning task-relevant and domain-invariant features. An influential approach involves incorporating gradient reversal layers to align feature spaces between domains without requiring explicit labels in the target domain. Techniques such as the domain-adversarial neural network \cite{ganin2016domain} exemplify this principle, achieving significant success in cross-domain applications \cite{sicilia2023domain}.

Loss reversal \cite{avram2024histnero} is a novel variation of the adversarial training paradigm that has demonstrated improved performance for domain-specific tasks. Instead of reversing the gradient at the input to the feature extractor, this method simply flips the sign of the domain discriminator's loss before optimization. As proposed in conjunction with the HistNERo dataset for historical NER in Romanian, loss reversal encourages the model to produce more distinct embeddings between domains, boosting performance on the task. Empirical results show that integrating loss reversal with pre-trained Romanian language models, namely BERT-ro \cite{dumitrescu2020birth} and DistilRoBERT \cite{avram2022distilling}, leads to substantial gains in strict F1-scores, with increases of over 10\% compared to the baseline non-adversarial approaches.

In addition to classical adversarial setups, recent research has explored integrating adversarial training with multitask learning and self-supervised pre-training \cite{smuadu2022legal,meftah2020multi}. These hybrid models aim to simultaneously optimize for multiple objectives, such as predicting token-level labels and aligning domain embeddings. For instance, topic-aware domain adaptation frameworks have incorporated latent topic modeling into adversarial setups, allowing models to learn domain-invariant features and topic-specific representations jointly. This combination has proven particularly effective for scientific keyphrase identification \cite{smuadu2023ta}, where incorporating adversarial examples further enhances performance. Such approaches demonstrate the versatility of adversarial techniques in addressing domain-specific challenges across a wide range of tasks.

Moreover, adversarial training has expanded beyond text classification and NER to address multilingual and cross-lingual tasks \cite{le2024lampat,zhang2024multilingual}. In these applications, adversarial discriminators are designed to promote language-invariant representations, enabling zero-shot or few-shot learning for low-resource languages. In addition, techniques such as lateral inhibition layers have been integrated with adversarial setups to enhance the separation of embeddings for specific tasks such as multi-word expression identification \cite{avram2023multilingual}. These advancements highlight the ongoing evolution of adversarial training, with each iteration introducing improved mechanisms to control the interaction between feature extraction and task-specific optimization.

\section{UniBERT}

\subsection{Training Data}

The UniBERT models were pre-trained on the latest dump of the Wikipedia corpus \cite{wikidump}, which was thoroughly filtered to ensure a high-quality dataset by employing the following heuristics:

\begin{itemize}
    \item Removed all the HTML artifacts left in the corpus using a simple regex.
    \item Removed all samples that contained only URLs or ISBNs.
    \item Removed all samples for which the count of digits represents more than 70\% of the length of the whole text.
    \item Removed all the hidden characters from the text.
    \item Removed all samples with less than five tokens using the multilingual BERT (mBERT) \cite{devlin-etal-2019-bert} tokenizer.
\end{itemize}

\begin{figure*}[!ht]
    \centering
    \includegraphics[width=0.99\linewidth]{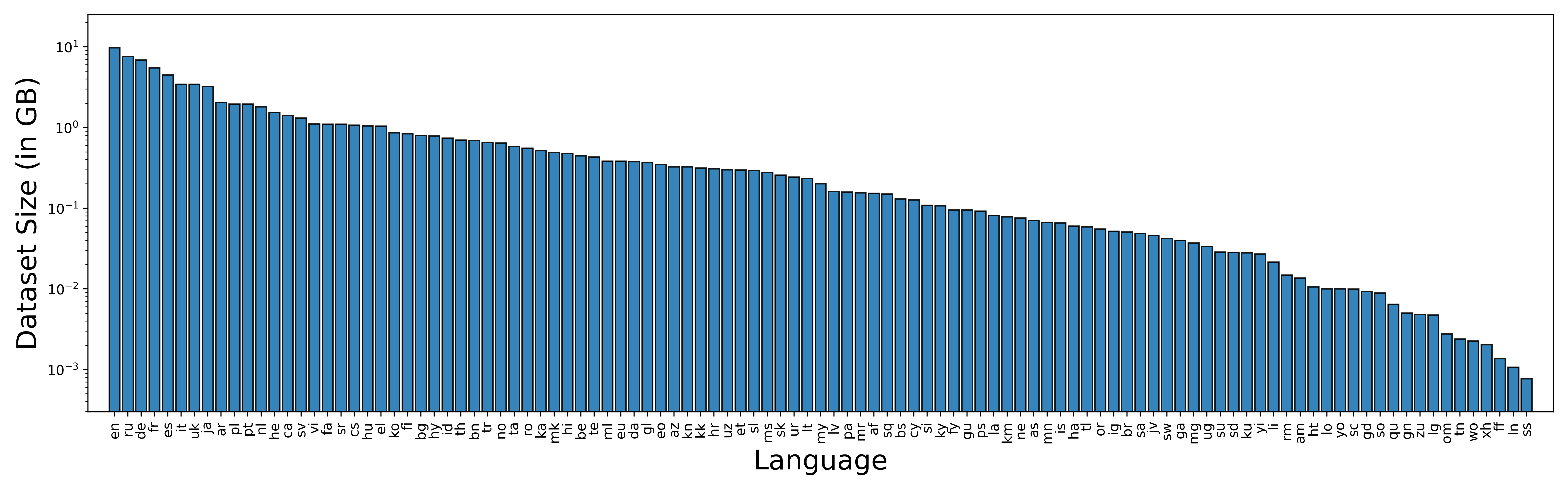}
    \caption{The amount of data in GB (log-scale) for each of the 107 languages that appear in the Wikipedia corpus was used to pre-train the UniBERT models.}
     \label{fig:langdistribution}   
\end{figure*}

The resulting dataset contained 82GB of textual data, with the English language being the most representative (i.e., 9.8GB of data) and the Swati language the least representative (i.e., 802K of data), as depicted in Figure \ref{fig:langdistribution}. We also measured the average number of tokens in the dataset using the mBERT tokenizer. We found that each sample in the dataset had approximately 110 tokens on average, with the maximum number of tokens being approximately 10K tokens\footnote{This number is truncated during pre-training and fine-tuning depending on the model's maximum context window.}.

\subsection{Model Architecture}

\begin{table*}
    \centering

    \caption{Comparison of our UniBERT models and the other multilingual BERT-based models available in the literature regarding their number of layers, number of hidden units in each layer, number of heads, vocabulary dimension, number of parameters, and model size.}
    
    \begin{tabular}{|l|r|r|r|r|r|r|}
         \toprule
         \textbf{Model} & \textbf{\#\ Layers} & \textbf{\#\ Hidden} & \textbf{\#\ Heads} & \textbf{Vocab Size} & \textbf{\#\ Params} & \textbf{Model Size} \\
         \midrule
         Distil-mBERT-base \cite{Sanh2019DistilBERTAD} & 6 & 768 & 12 & 120K & 135M & 543M \\
         mBERT-base \cite{devlin-etal-2019-bert} & 12 & 768 & 12 & 120K & 179M & 715M \\
         XLM-RoBERTa-base \cite{conneau2019cross} & 12 & 768 & 12 & 251K & 279M & 1.12G \\
         XLM-RoBERTa-large \cite{conneau2019cross} & 24 & 1024 & 16 & 251K & 560M & 2.24G \\
         XLM-V-base \cite{liang2023xlm} & 12 & 768 & 12 & 901K & 778M & 3.12G \\
         \midrule
         \textit{UniBERT-Small} (ours) & 6 & 512 & 8 & 120K & 87M & 334M \\
         \textit{UniBERT-XSmall} (ours) & 4 & 256 & 8 & 120K & 39M & 148M \\
         \textit{UniBERT-XXSmall} (ours) & 4 & 128 & 4 & 120K & 19M & 74M \\
         
         \bottomrule
    \end{tabular}
    \label{tab:model_size}
\end{table*}

In this work, we introduce three distinct UniBERT models: UniBERT-Small, UniBERT-XSmall, and UniBERT-XXSmall. These models provide varied configurations to address different computational needs and efficiency requirements. Each variant is designed to minimize the number of parameters and the overall size of the model, allowing efficient multilingual processing without sacrificing the core model performances.

Unlike prominent multilingual models such as Distil-mBERT-base \cite{Sanh2019DistilBERTAD}, mBERT-base \cite{devlin-etal-2019-bert}, and the XLM-RoBERTa series \cite{conneau2019cross}, our UniBERT models offer a streamlined architecture. Table \ref{tab:model_size} shows that the UniBERT-Small model includes 6 layers with 512 hidden units per layer and 8 attention heads, which make it comparable in layer count to Distil-mBERT-base but with fewer hidden units and heads, resulting in a reduced number of parameters (87M versus 135M) and a smaller model size (334M versus 543M). This reduction is achieved while maintaining a similar vocabulary dimension of 120K, indicating that UniBERT-Small retains sufficient linguistic flexibility for multilingual tasks.

Further comparison of our UniBERT models with the XLM family reveals significant efficiency differences. While XLM-RoBERTa-base (i.e., 279M parameters, 1.12G size), XLM-RoBERTa-large (i.e., 560M parameters, 2.24G size), and XLM-V-base (i.e., 778M parameters, 3.12G size) \cite{liang2023xlm} rely on increasingly larger architectures with extensive vocabularies, UniBERT-Small reduces the model size to less than 12\% of XLM-V-base's footprint. This efficiency gap widens even more with UniBERT-XSmall (39M parameters) and UniBERT-XXSmall (19M parameters), the latter being approximately 30 times smaller than XLM-RoBERTa-large and 40 times smaller than XLM-V-base, while maintaining effective multilingual capabilities. These dramatic size reductions demonstrate our approach's effectiveness in creating compact yet functional models for resource-constrained environments.

\subsection{Training Process}

\begin{figure*}[!ht]
    \centering
    \includegraphics[width=0.90\linewidth]{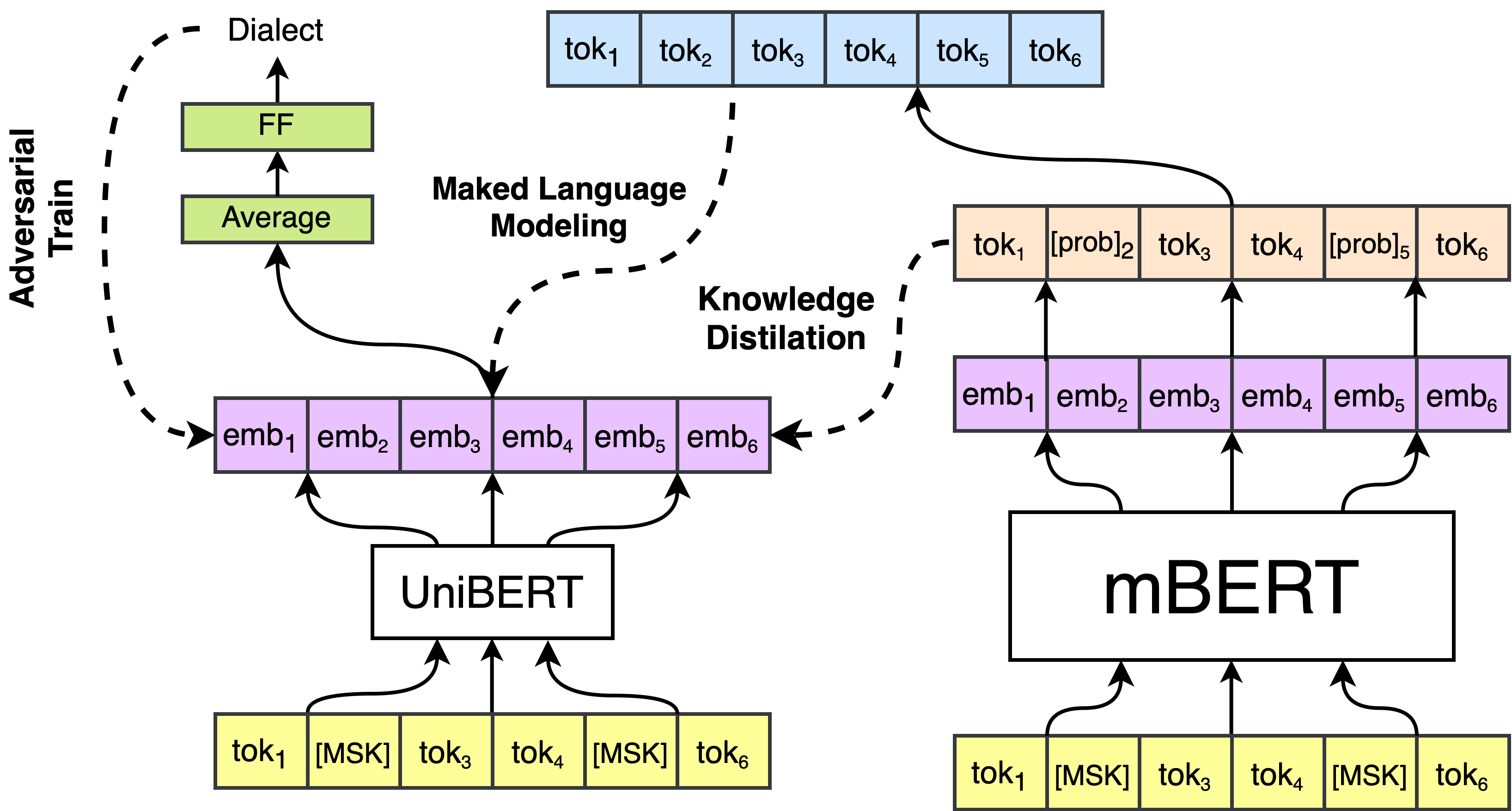}
    \caption{UniBERT pre-training methodology. Given a sequence of tokens, both UniBERT and mBERT produce a corresponding set of embeddings. The UniBERT embeddings are first used in the masked language modeling task, as in the original mBERT pre-training. Then, those embeddings are averaged out, and a feed-forward layer is used to predict the language of the input sentence. The learning signal from this classifier is used to train the UniBERT model to produce language-independent embeddings adversarially. Finally, the knowledge distillation technique is used to align the probability distributions of the two models.}
    \label{fig:archsystem}
\end{figure*}

Figure \ref{fig:archsystem} shows the general pre-training methodology. The training framework involves a novel approach to train a smaller, more efficient BERT-based language model, called UniBERT, through three main techniques: (1) masked language modeling as for the mBERT model, (2) adversarial training using a language discriminator \cite{chen2018adversarial} with the loss reversal technique \cite{avram2024histnero}, and (3) knowledge distillation with the mBERT model serving as the teacher model.

\subsubsection{Masked Language Modeling}

Masked language modeling (MLM) \cite{devlin-etal-2019-bert} is a core component of BERT-based models that enables them to learn bidirectional representations of language. In MLM, a percentage of tokens within the input sequence are masked randomly, and the model is trained to predict these masked tokens based on the surrounding context. Formally, let $x=(x_1, x_2, x_3, ..., x_n)$ be an input sequence of tokens, where $x_i$ represents each token. A subset of these tokens, called $x_{masked} \subset x$, is replaced by a special token \texttt{[MASK]}, and the model optimizes the probability of recovering the original tokens $P(x_{masked}|x_{\setminus masked})$ given the context. This choice can be achieved by minimizing the cross-entropy loss over the masked tokens, as shown in Equation \ref{eq:mlm}:

\begin{equation}
    \mathcal{L}_{\text{MLM}} = - \sum_{x_{\text{masked}}} \log P(x_{\text{masked}} | x_{\setminus \text{masked}}; \theta_{\text{student}})
    \label{eq:mlm}
\end{equation}
where $\theta_{\text{student}}$ represents the parameters of the student model. The MLM objective enables the model to capture contextual dependencies on both sides of a token, effectively learning a bidirectional understanding of language structure. In UniBERT, MLM serves as a foundational pre-training task, helping the model learn syntactic and semantic representations that are robust and generalizable across different applications.

\subsubsection{Adversarial Training}

In UniBERT, adversarial training \cite{chen2018adversarial} involves training a language classifier that learns to distinguish between different language inputs, using the average of the embeddings $\bar{e_i}$ generated by the student model. To encourage language invariance, a loss reversal technique is applied, which involves simply reversing the sign of the loss $\mathcal{L}_{\text{disc}}$ provided by the language discriminator, as shown in Equation \ref{eq:adv}:

\begin{equation}
    y_i = FF(\bar{e_i})
\end{equation}

\begin{equation}
    \mathcal{L}_{\text{adv}} = - \sum_{i} y_i \log P(y_i | \bar{e_i}; \theta_{\text{disc}})
    \label{eq:adv}
\end{equation}
where $FF$ is the feed-forward layer that produce the language probabilities $y_i$ and $\theta_{\text{disc}}$ are the parameters of the language discriminator model.

\subsubsection{Knowledge Distillation}

Knowledge distillation \cite{hinton2015distilling} is a model compression technique in which a smaller model (known as the student) learns from a larger, pre-trained model (known as the teacher) by approximating its predictions. In UniBERT, knowledge distillation is implemented by using mBERT as the teacher model, allowing UniBERT to inherit multilingual capabilities from mBERT while maintaining a more compact and efficient structure. The objective of knowledge distillation is to minimize the Kullback-Leibler (KL) divergence between the teacher and student model outputs,  as shown in Equation \ref{eq:kd}:

\begin{equation}
    \mathcal{L}_{\text{KD}} = \text{KL}(z_{\text{teacher}} \, || \, z_{\text{student}}) = - \sum_i z_{\text{teacher}}^{(i)} \log \frac{z_{\text{teacher}}^{(i)}}{z_{\text{student}}^{(i)}}
    \label{eq:kd}
\end{equation}
where $z_{teacher}$ and $z_{student}$ represent the logits of the teacher and student models. This objective aligns the output distribution of the student model with that of the teacher, effectively transferring knowledge without requiring the student to have the same computational complexity as the teacher.

\subsubsection{Training Objective}

The general training objective for UniBERT combines the objectives of MLM, adversarial training, and knowledge distillation, represented as the weighted sum of the three individual losses, as shown in Equation \ref{eq:total_loss}:

\begin{equation}
    \mathcal{L}_{\text{total}} = \lambda_{\text{MLM}}\mathcal{L}_{\text{MLM}} - \lambda_{\text{adv}}\mathcal{L}_{\text{adv}} + \lambda_{\text{KD}}\mathcal{L}_{\text{KD}}
    \label{eq:total_loss}
\end{equation}
where $\mathcal{L}_{\text{MLM}}$, $\mathcal{L}_{\text{adv}}$, $\mathcal{L}_{\text{KD}}$ are the objectives of MLM, adversarial, and knowledge distillation, respectively, whereas $\lambda_{\text{MLM}}$, $\lambda_{\text{adv}}$, and $\lambda_{\text{KD}}$ are their corresponding weights. The MLM loss $\mathcal{L}_{\text{MLM}}$ enables UniBERT to learn semantic representations by predicting masked tokens from context. In contrast, the adversarial loss $\mathcal{L}_{\text{adv}}$ encourages the model to be language-agnostic by reversing the gradient of the language classifier. Finally, the knowledge distillation loss $\mathcal{L}_{\text{KD}}$ allows UniBERT to mimic a larger multilingual BERT model, preserving multilingual capabilities in a more compact form. Together, these objectives ensure that UniBERT is both efficient and capable of handling cross-lingual tasks with a high degree of generalization.

\subsection{Training Hyperparameters}

The UniBERT models were pre-trained for a total of 1M steps with a batch size of 128, utilizing a learning rate of 5e-4 with a linear scheduling approach and an initial warmup period spanning the first 50,000 steps. We selected the AdamW optimizer \cite{loshchilov2018decoupled} to optimize the stability of the training. In addition, gradient clipping \cite{pmlr-v28-pascanu13} was applied with a maximum gradient norm set to 5, effectively preventing large gradient updates. A weight decay of 0.01 was used to focus on optimization and improve the primary parameters without penalizing their magnitudes over time.

We set the training objective coefficients for the MLM loss $\lambda_{\text{MLM}}$ to 0.5, for the knowledge distillation loss $\lambda_{\text{KD}}$ to 0.4, and for the adversarial loss $\lambda_{\text{adv}}$ to 0.1. Furthermore, to refine UniBERT’s language understanding for the MLM objective, we randomly selected 15\% of tokens for masking, the probability being applied uniformly, ensuring an even distribution between tokens. The MLM loss incorporated label smoothing, set to a factor of 0.7, which helped alleviate overconfidence in predictions by distributing some probability mass across non-target classes. Finally, a softmax temperature of 2 was utilized to adjust the output distribution's sharpness, effectively balancing the confidence of the UniBERT models across potential predictions. We trained each UniBERT model on two NVIDIA A100 with 40GB of GPU memory and 16 CPU cores for 30 days.

\section{Evaluation Datasets}

To evaluate the UniBERT models, we considered two scenarios: (1) "each" - measuring their performance by training and evaluating each of the individual languages in the dataset, and (2) "all" - training the model in all languages and calculating the average performance in each language. Experiments were carried out on four key NLP tasks as follows:

\begin{itemize}
    \item \textbf{Named Entity Recognition (NER)}, whose goal is to identify and categorize specific entities within the text, such as people, organizations, and locations. We evaluated UniBERT on the CoNLL-2002 \cite{tjong-kim-sang-2002-introduction} (i.e., for the Spanish and Dutch languages) and CoNLL-2003 \cite{tjong-kim-sang-de-meulder-2003-introduction} (i.e., the English language\footnote{The German subset is not available for public usage, so we did not evaluate the performance of the UniBERT models on this language.}) datasets, which contain entities labeled as PER (person), ORG (organization), LOC (location), and MISC (miscellaneous). The CoNLL-2002 dataset contains approximately 12K sentences in Spanish and 24K sentences in Dutch, while CoNLL-2003 includes around 21K sentences in English. Performance was measured using the F1-score, which balances precision and recall for entity categorization.
    
    \item \textbf{Natural Language Inference (NLI)}, whose goal is to assess the model's ability to understand relationships between pairs of sentences, classifying each as entailment, contradiction, or neutral. We used the Cross-lingual Natural Language Inference (XNLI) dataset \cite{conneau2018xnli}, which includes about 400K sentence pairs for each of the 15 available languages. We used the F1-score as the primary evaluation metric, which balances precision and recall for sentence classification.
    
    \item \textbf{Question Answering (QA}, which measures the ability of the models to provide answers based on a provided context. We evaluated UniBERT models on the MultiLingual Question Answering (MLQA) dataset \cite{lewis2020mlqa}, which consists of more than 90K context-question-answer triplets in seven languages. Each input includes a context paragraph and a question, with the goal of identifying the correct text span as the answer. We measured performance using F1-scores for partial matches.
    
    \item \textbf{Semantic Textual Similarity (STS)}, which measures the degree of semantic similarity between sentence pairs, rated on a scale from 0 (no similarity) to 5 (high similarity). We used the SemEval-2022 Task 8 (STS22) dataset \cite{chen2022semeval}, which provides over 9K sentence pairs with human-annotated similarity scores in ten languages. Additionally, because this dataset contains samples with the two sentences in different languages, we differentiate between fine-tuning the model on the whole dataset, which includes the samples in different languages (case called  "all" in this paper) and fine-tuning the model only on the samples that contain sentences in the same language (case called "mono" in this paper). We evaluated UniBERT performance using the Pearson correlation coefficient \cite{Rodgers01021988}, reflecting the alignment between predicted scores and human judgments.
\end{itemize}

Furthermore, we compared the performance of the UniBERT models in these tasks with well-known baselines in the literature: mBERT-base, Distil-mBERT-base, and XLM-RoBERTa (base and large variants).

\section{Results}

\subsection{Named Entity Recognition}

Table \ref{tab:ner_results} summarizes the F1-scores achieved on the CoNLL-2002 and CoNLL-2003 datasets. The results are somewhat mixed. Among the baseline models, XLM-RoBERTa-large achieves the highest performance when individually fine-tuned, reaching F1-scores of 92.70\% (en), 90.05\% (es), and 93.35\% (nl), for an overall average of 92.03\%. In contrast, when trained jointly in all languages, its performance drops slightly (i.e., to 91.82\%, 88.45\%, and 91.77\%, for an average of 90.67\%). Similarly, mBERT-base, XLM-RoBERTa-base, and XLM-V-base show marginally lower scores in the "all" configuration than their "each" counterparts, with the latter obtaining the highest F1-score in the "all" training configuration.

The proposed UniBERT models, while inherently more compact, display an interesting trend. For instance, UniBERT-Small achieves an average F1-score of 83.25\% when individually fine-tuned, but this improves to 84.40\% under multilingual joint  training. A similar improvement is observed for the even smaller UniBERT-XSmall and UniBERT-XXSmall models, with average scores increasing from 77.25\% to 78.80\% and from 71.64\% to 74.00\%, respectively. These gains suggest that the joint training strategy effectively mitigates some of the capacity limitations inherent to more compact architectures, particularly for non-English languages.

\begin{table}
    \centering

    \caption{Results of NER on the CoNLL-2002 and CoNLL-2003 datasets. We report the F1-score of the models that were fine-tuned for each language (i.e., case "each") and for all languages at the same time (i.e., case "all"), evaluated on each language, together with the average F1-score.}
    
    \begin{tabular}{|l|c|c|c|c|c|}
         \toprule
         \textbf{Model} & \textbf{Train} & \textbf{en} & \textbf{es} & \textbf{nl} & \textbf{Avg} \\
         \midrule
         \multirow{ 2}{*}{Distil-mBERT-base \cite{Sanh2019DistilBERTAD}}  & each & 89.26 & 86.50 & 87.02 & 87.59 \\
          & all & 88.86 & 86.67 & 86.61 & 87.38 \\
         \midrule
         \multirow{ 2}{*}{mBERT-base \cite{devlin-etal-2019-bert}} & each & 90.39 & 88.48 & 90.05 & 89.96 \\
          & all & 90.09 & 87.99 & 89.64 & 89.24 \\
          \midrule
         \multirow{ 2}{*}{XLM-RoBERTa-base \cite{conneau2019cross}} & each & 91.02 & 88.85 & 91.31 & 90.39 \\
          & all & 91.40 & 88.90 & 90.33 & 90.21 \\
          \midrule
         \multirow{ 2}{*}{XLM-RoBERTa-large \cite{conneau2019cross}} & each & 92.70 & 90.05 & 93.35 & 92.03 \\
          & all & 91.82 & 88.45 & 91.77 &  90.67 \\
          \midrule
          \multirow{ 2}{*}{XLM-V-base \cite{liang2023xlm}} & each & 92.41 & 88.95 & 94.23 & 91.83 \\
 & all & 90.96 & 89.62 & 93.51 & 91.36 \\
          \midrule
         \multirow{ 2}{*}{\textit{UniBERT-Small} (ours)} & each & 86.21 & 81.12 & 83.25 & 83.25 \\
          & all & 86.90 & 83.23 & 83.09 & 84.40 \\
          \midrule
         \multirow{ 2}{*}{\textit{UniBERT-XSmall} (ours)} & each & 81.74 & 77.78 & 72.25 & 77.25 \\
          & all & 82.83 & 78.47 & 75.10 & 78.80 \\
          \midrule
         \multirow{ 2}{*}{\textit{UniBERT-XXSmall} (ours)} & each & 78.32 & 69.16 & 67.45 & 71.64 \\
          & all & 80.14 & 73.14 & 69.90 & 74.00 \\
         
         \bottomrule
    \end{tabular}
    \label{tab:ner_results}
\end{table}

\subsection{Natural Language Inference}

Table \ref{tab:nli_results} presents the F1-scores on the XNLI dataset. The results are somewhat mixed on the NLI task, with the larger models leading overall performance. For instance, XLM-RoBERTa-large achieves an average F1-score of 82.29\% when fine-tuned independently in each language, which further increases to 83.30\% in multilingual joint  training. Likewise, XLM-RoBERTa-base shows a notable improvement from 75.18\% to 79.04\% when switching from individual to joint training. In contrast, mBERT-base exhibits a moderate increase of an average of 72.64\% to 73.37\%. Also, Distil-mBERT-base remains relatively stable with averages of 67.68\% (i.e., case "each") and 67.74\% (i.e., case "all"). Finally, XLM-V-base obtained the highest performance in the "all" training configuration with an F1-score of 83.92\%, its "each" training performance being slightly lower than XLM-RoBERTa-large (i.e., an F1-score of 81.92\%).

Our UniBERT variants, although more compact, benefit appreciably from the joint training strategy. The average F1-score of UniBERT-Small increases from 61.52\% in the "each" setting to 65.69\% when all languages are considered at once. The smaller UniBERT-XSmall and UniBERT-XXSmall models experience even more pronounced improvements, with their average F1-scores rising from 54.36\% to 61.28\%, and from 48.07\% to 54.45\%, respectively. These gains, which range roughly between 4\% and 7\%, suggest that sharing cross-lingual representations during fine-tuning is especially beneficial for smaller models, helping compensate for their reduced capacity.

\begin{table*}
    \centering

    \caption{Results on XNLI. We report the F1-score of the models that were fine-tuned for each language (i.e., case "each") and for all languages at the same time (i.e., case "all"), evaluated on each language, together with the average F1-score.}
    \resizebox{\textwidth}{!}{
    \begin{tabular}{|l|c|c|c|c|c|c|c|c|c|c|c|c|c|c|c|c|c|}
         \toprule
         \textbf{Model} & \textbf{Train} & \textbf{ar} & \textbf{bg} & \textbf{de} & \textbf{el} & \textbf{en} & \textbf{es} & \textbf{fr} & \textbf{hi} & \textbf{ru} & \textbf{sw} & \textbf{th} & \textbf{tr} & \textbf{ur} & \textbf{vi} & \textbf{zh} & \textbf{Avg}  \\
         \midrule
         \multirow{ 2}{*}{Distil-mBERT-base \cite{Sanh2019DistilBERTAD}} & each & 67.72 & 70.21 & 72.10 & 69.59 & 76.91 & 72.76 & 72.43 & 47.93 & 69.60 & 61.75 & 62.29 & 67.39 & 61.19 & 70.10 & 73.37 & 67.68 \\
          & all & 65.13 & 69.92 & 70.76 & 68.06 & 76.33 & 73.15 & 72.97 & 62.45 & 68.71 & 59.16 & 59.77 & 66.26 & 62.72 & 69.54 & 71.23 & 67.74 \\
          \midrule
         \multirow{ 2}{*}{mBERT-base \cite{devlin-etal-2019-bert}} & each & 70.29 & 75.11 & 75.93 & 74.17 & 82.08 & 77.40 & 77.02 & 66.77 & 74.07 & 64.66 & 66.03 & 71.24 & 63.49 & 74.87 & 76.57 & 72.64 \\
          & all & 71.89 & 76.39 & 76.65 & 74.45 & 81.76 & 77.61 & 77.08 & 68.58 & 75.03 & 64.20 & 66.47 & 72.39 & 65.97 & 76.04 & 76.09 & 73.37 \\
          \midrule
         \multirow{ 2}{*}{XLM-RoBERTa-base \cite{conneau2019cross}} & each & 73.63 & 78.65 & 75.68 & 75.11 & 83.55 & 79.25 & 77.73 & 71.10 & 76.53 & 68.36 & 75.01 & 74.87 & 65.50 & 76.53 & 76.31 & 75.18 \\
          & all & 76.78 & 81.57 & 80.59 & 80.12 & 85.16 & 82.13 & 80.79 & 80.79 & 79.34 & 71.60 & 77.81 & 77.84 & 72.72 & 79.24 & 79.16 & 79.04 \\
          \midrule
         \multirow{ 2}{*}{XLM-RoBERTa-large \cite{conneau2019cross}} & each & 82.07 & 85.48 & 84.61 & 84.05 & 88.71 & 85.57 & 84.66 & 79.61 & 83.01 & 76.36 & 81.22 & 81.49 & 72.38 & 82.80 & 82.45 & 82.29 \\
          & all & 82.84 & 85.88 & 85.12 & 84.97 & 88.92 & 85.91 & 85.05 & 81.42 & 83.18 & 77.79 & 81.37 & 82.41 & 77.46 & 83.65 & 83.40 & 83.30 \\
          \midrule
          \multirow{ 2}{*}{XLM-V-base \cite{liang2023xlm}} & each & 81.50 & 84.55 & 84.71 & 83.20 & 87.85 & 85.65 & 84.75 & 80.03 & 82.16 & 76.63 & 81.37 & 80.60 & 70.51 & 82.94 & 82.43 & 81.92 \\
 & all & 84.12 & 86.38 & 85.27 & 85.19 & 89.16 & 86.28 & 85.99 & 81.22 & 83.31 & 78.96 & 81.84 & 84.59 & 78.60 & 83.51 & 84.39 & 83.92 \\
         \midrule
         \multirow{ 2}{*}{\textit{UniBERT-Small} (ours)} & each & 59.06 & 61.91 & 63.63 & 61.76 & 68.33 & 64.97 & 65.20 & 58.22 & 61.10 & 56.25 & 57.89 & 60.71 & 55.81 & 64.46 & 63.52 & 61.52 \\
         & all & 63.46 & 67.17 & 68.56 & 66.34 & 74.08 & 70.88 & 70.19 & 61.57 & 66.00 & 58.30 & 57.71 & 64.08 & 59.96 & 69.03 & 68.01 & 65.69 \\
          \midrule
         \multirow{ 2}{*}{\textit{UniBERT-XSmall} (ours)} & each & 54.61 & 52.89 & 59.62 & 53.51 & 59.54 & 55.56 & 56.04 & 53.16 & 56.04 & 49.23 & 49.46 & 52.07 & 50.40 & 56.12 & 57.16 & 54.36 \\
         & all & 60.12 & 61.92 & 63.46 & 61.76 & 67.93 & 64.45 & 65.23 & 58.61 & 60.99 & 55.48 & 55.23 & 60.42 & 56.42 & 64.25 & 63.00 & 61.28 \\
          \midrule
         \multirow{ 2}{*}{\textit{UniBERT-XXSmall} (ours)} & each & 46.89 & 48.89 & 50.13 & 47.46 & 52.84 & 50.42 & 48.38 & 47.02 & 48.63 & 42.42 & 46.05 & 46.25 & 47.12 & 50.07 & 48.48 & 48.07 \\
         & all & 52.61 & 53.91 & 56.71 & 55.34 & 59.90 & 57.00 & 56.93 & 53.07 & 56.87 & 49.36 & 47.88 & 53.47 & 51.88 & 56.72 & 55.13 & 54.45 \\
         \bottomrule
    \end{tabular}
    }
    \label{tab:nli_results}
\end{table*}

\subsection{Question Answering}

Table \ref{tab:qa_results} reports the F1-scores on the MLQA dataset for each model. The baseline models consistently outperform in absolute terms. For example, XLM-RoBERTa-large achieves an average F1-score of 63.51\% when individually fine-tuned, which further improves to 64.85\% under joint training. Similarly, XLM-RoBERTa-base records average scores of 58.93\% (i.e., case "each") and 59.97\% (i.e., case "all"), mBERT-base reaches 57.86\% (i.e., case "each") and 59.21\%  (i.e., case "all"), while Distil-mBERT-base achieves 51.95\%(i.e., case "each") and 55.21\%  (i.e., case "all"). XLM-V-base performs comparably to XLM-RoBERTa-large, with scores of 63.34\% (i.e., case "each") and 65.57\% (i.e., case "all"), showing the strongest improvement among larger models on this task.

Our proposed UniBERT models, despite their reduced parameter count, benefit noticeably from multilingual joint  fine-tuning. UniBERT-Small, for example, sees its average F1-score rise from 42.87\% to 44.59\%. The improvements are even more pronounced for smaller variants: UniBERT-XSmall increases from 28.43\% to 36.53\%, and UniBERT-XXSmall from 17.57\% to 21.33\%. These enhancements, which range from approximately 2\% to more than 8\%, highlight the advantages of using cross-lingual data during training, particularly for models with limited capacity.

\begin{table*}
    \centering

    \caption{Results on MLQA. We report the F1-score of the models that were fine-tuned for each language (i.e., case "each") and for all languages at the same time (i.e., case "all"), evaluated on each language, together with the average F1-score.}
    
    \begin{tabular}{|l|c|c|c|c|c|c|c|c|c|}
         \toprule
         \textbf{Model} & \textbf{Train} & \textbf{ar} & \textbf{de} & \textbf{en}  & \textbf{es} & \textbf{hi} & \textbf{vi} & \textbf{zh} &  \textbf{Avg} \\
         \midrule
         \multirow{ 2}{*}{Distil-mBERT-base \cite{Sanh2019DistilBERTAD}} & each & 43.28 & 54.32 & 73.42 & 59.68 & 48.90 & 56.07 & 28.04 & 51.95 \\
          & all & 46.91 & 58.23 & 75.36 & 63.94 & 52.72 & 60.27 & 29.07 & 55.21 \\
         \midrule
         \multirow{ 2}{*}{mBERT-base \cite{devlin-etal-2019-bert}} & each & 51.43 & 59.41 & 78.26 & 65.71 & 56.03 & 63.42 & 30.82 & 57.86 \\
          & all & 51.99 & 61.76 & 78.59 & 67.25 & 57.95 & 65.08 & 31.88 & 59.21 \\
          \midrule
         \multirow{ 2}{*}{XLM-RoBERTa-base \cite{conneau2019cross}} & each & 52.09 & 59.48 & 77.38 & 64.75 & 61.48 & 65.75 & 31.59 & 58.93 \\
          & all & 53.59 & 61.08 & 78.06 & 66.36 & 63.05 & 66.85 & 30.80 & 59.97 \\
          \midrule
         \multirow{ 2}{*}{XLM-RoBERTa-large \cite{conneau2019cross}} & each & 58.02 & 65.61 & 80.65 & 70.46 & 68.02 & 69.62 & 32.24 & 63.51 \\
          & all & 59.37 & 67.41 & 81.88 & 71.95 & 69.85 & 71.05 & 32.50 & 64.85 \\
          \midrule
          \multirow{ 2}{*}{XLM-V-base \cite{liang2023xlm}} & each & 58.33 & 65.81 & 80.89 & 70.66 & 66.30 & 68.81 & 32.55 & 63.34 \\
 & all & 59.90 & 66.57 & 83.31 & 71.44 & 69.15 & 70.65 & 38.03 & 65.57 \\      
         \midrule
         \multirow{2}{*}{\textit{UniBERT-Small} (ours)} & each & 32.29 & 43.96 & 63.05 & 51.14 & 42.30 & 49.87 & 17.49 & 42.87 \\
         & all & 36.76 & 46.18 & 64.41 & 52.42 & 42.02 & 51.05 & 19.31 & 44.59 \\
         \midrule
         \multirow{2}{*}{\textit{UniBERT-XSmall} (ours)} & each & 16.18 & 27.24 & 51.69 & 39.64 & 12.19 & 38.13 & 13.97 & 28.43 \\
         & all & 29.86 & 39.66 & 51.66 & 42.52 & 34.95 & 42.09 & 14.97 & 36.53 \\
         \midrule
         \multirow{ 2}{*}{\textit{UniBERT-XXSmall} (ours)} & each & 10.12 & 20.43 & 25.21 & 23.66 & 17.20 & 20.05 & 6.32 & 17.57 \\
         & all & 18.04 & 26.00 & 28.27 & 26.41 & 18.60 & 24.11 & 7.86 & 21.33 \\
         
         \bottomrule
    \end{tabular}
    \label{tab:qa_results}
\end{table*}

\subsection{Semantic Textual Similarity}

Table \ref{tab:sts_results} presents the Pearson correlation coefficients obtained on the STS22 dataset. In this evaluation, we consider three training configurations: fine-tuning on each language individually (i.e., case "each"), on all the monolingual subsets simultaneously (i.e., case "mono"), and on the entire dataset that includes cross-lingual samples (i.e., case "all"). Note that results are provided only for the "all" and "mono" settings for Chinese, Russian, and Italian (i.e., languages for which just a test set is available).

Among the baseline models, mBERT-base achieves an average coefficient of 0.674 when fine-tuned on each language individually. Its performance improves to 0.723 with monolingual training and settles to 0.710 when including cross-lingual data. Similarly, XLM-RoBERTa-large obtains average scores of 0.628, 0.737, and 0.743 in the "each", "mono", and "all" settings, respectively, suggesting that joint training, especially when leveraging cross-lingual samples, can boost semantic similarity performance. XLM-V-base follows a similar pattern with scores of 0.621, 0.729, and 0.730 in the respective settings.

In contrast, the UniBERT variants, while more compact, show lower overall correlations. For example, UniBERT-Small records an average of 0.547 in the "each" configuration, increasing to 0.614 with monolingual training and 0.624 when cross-lingual examples are included. UniBERT-XSmall and UniBERT-XXSmall follow a similar pattern, with their average coefficients increasing by approximately 0.07–0.08 points when moving from individual to joint training strategies.

\begin{table*}
    \centering

    \caption{Results on the STS22 dataset. We report the Pearson correlation coefficients of the models fine-tuned on each language (i.e., case "each"), on all the monolingual subsets at the same time (i.e., case "mono"), and on the whole dataset, which includes cross-lingual samples (i.e., case "all"), evaluated on each language, together with the average Pearson correlation coefficient. The Italian, Russian, and Chinese languages provide only the test set, so we do not give any results on the "each" category for them.}

   \resizebox{\textwidth}{!}{
   \begin{tabular}{|l|c|c|c|c|c|c|c|c|c|c|c|c|}
         \toprule
         \textbf{Model} & \textbf{Train} & \textbf{ar} & \textbf{de} & \textbf{en}  & \textbf{es} & \textbf{fr} & \textbf{it} & \textbf{pl} & \textbf{ru} & \textbf{tr} & \textbf{zh} & \textbf{Avg} \\
         \midrule
         \multirow{3}{*}{Distil-mBERT-base \cite{Sanh2019DistilBERTAD}} & each & 0.521 & 0.578 & 0.719 & 0.681 & 0.592 & - & 0.461 & - & 0.569 & - & 0.589 \\
         & mono & 0.675 & 0.710 & 0.718 & 0.751 & 0.820 & 0.713 & 0.536 & 0.654 & 0.685 & 0.715 & 0.698 \\
          & all & 0.637 & 0.600 & 0.715 & 0.723 & 0.789 & 0.659 & 0.519 & 0.584 & 0.583 & 0.684 & 0.649 \\
         \midrule
         \multirow{3}{*}{mBERT-base \cite{devlin-etal-2019-bert}} & each & 0.710 & 0.668 & 0.729 & 0.741 & 0.641 & - & 0.572 & - & 0.655 & - & 0.674 \\
         & mono & 0.720 & 0.711 & 0.722 & 0.772 & 0.842 & 0.745 & 0.642 & 0.672 & 0.694 & 0.710 & 0.723 \\
          & all & 0.718 & 0.659 & 0.733 & 0.760 & 0.826 & 0.717 & 0.626 & 0.685 & 0.652 & 0.722 & 0.710 \\
          \midrule
         \multirow{3}{*}{XLM-RoBERTa-base \cite{conneau2019cross}} & each & 0.582 & 0.602 & 0.732 & 0.713 & 0.658 & - & 0.455 & - & 0.639 & - & 0.626 \\
         & mono & 0.802 & 0.690 & 0.730 & 0.739 & 0.833 & 0.723 & 0.565 & 0.631 & 0.656 & 0.723 & 0.709 \\
          & all & 0.757 & 0.693 & 0.737 & 0.770 & 0.829 & 0.749 & 0.514 & 0.647  & 0.677 & 0.714 & 0.709 \\
          \midrule
         \multirow{3}{*}{XLM-RoBERTa-large \cite{conneau2019cross}} & each & 0.592 & 0.716 & 0.743 & 0.704 & 0.506 & - & 0.577 & - & 0.558 & - & 0.628 \\
         & mono & 0.756 & 0.763 & 0.761 & 0.756 & 0.837 & 0.770 & 0.591 & 0.671 & 0.712 & 0.752 & 0.737 \\
          & all & 0.774 & 0.758 & 0.745 & 0.761 & 0.855 & 0.778 & 0.602 & 0.674 & 0.733 & 0.753 & 0.743 \\
          \midrule
          \multirow{3}{*}{XLM-V-base \cite{liang2023xlm}} & each & 0.555 & 0.708 & 0.746 & 0.708 & 0.505 & - & 0.568 & - & 0.562 & - & 0.621 \\
& mono & 0.750 & 0.763 & 0.723 & 0.746 & 0.831 & 0.743 & 0.597 & 0.674 & 0.719 & 0.751 & 0.729 \\
 & all & 0.756 & 0.761 & 0.747 & 0.724 & 0.832 & 0.781 & 0.615 & 0.618 & 0.725 & 0.747 & 0.730 \\
         \midrule
         \multirow{3}{*}{\textit{UniBERT-Small} (ours)} & each & 0.436 & 0.665 & 0.629 & 0.556 & 0.601 & - & 0.483 & - & 0.427 & - & 0.547 \\
         & mono & 0.499 & 0.686 & 0.670 & 0.591 & 0.673 & 0.718 & 0.535 & 0.639 & 0.516 & 0.622 & 0.614 \\
          & all & 0.515 & 0.701 & 0.683 & 0.577 & 0.680 & 0.723 & 0.540 & 0.630 & 0.570 & 0.625 & 0.624 \\
          \midrule
         \multirow{3}{*}{\textit{UniBERT-XSmall} (ours)} & each & 0.419 & 0.638 & 0.604 & 0.534 & 0.577 & - & 0.464 & - & 0.410 & - & 0.520 \\
         & mono & 0.479 & 0.659 & 0.643 & 0.567 & 0.646 & 0.689 & 0.514 & 0.613 & 0.495 & 0.596 & 0.590 \\
         & all  & 0.494 & 0.673 & 0.656 & 0.554 & 0.653 & 0.694 & 0.518 & 0.605 & 0.547 & 0.600 & 0.600 \\
         \midrule
         \multirow{3}{*}{\textit{UniBERT-XXSmall} (ours)} & each & 0.401 & 0.612 & 0.579 & 0.512 & 0.553 & - & 0.444 & - & 0.393 & - & 0.500 \\
         & mono & 0.459 & 0.631 & 0.616 & 0.544 & 0.619 & 0.661 & 0.492 & 0.588 & 0.475 & 0.572 & 0.565 \\
         & all  & 0.474 & 0.645 & 0.628 & 0.531 & 0.626 & 0.665 & 0.497 & 0.580 & 0.524 & 0.575 & 0.574 \\
         
         \bottomrule
    \end{tabular}
    }
    \label{tab:sts_results}
\end{table*}

\subsection{Out-of-Domain Analysis}

We further performed an out-of-domain analysis of the UniBERT models for the legal domain using the MAPA dataset \cite{DeGibertBonet2022}. MAPA is a NER dataset designed for anonymization tasks in legal documents across 22 European languages. The dataset contains annotated entities that require anonymization in official documents, such as PERSON, ORGANIZATION, LOCATION, and other domain-specific categories. This evaluation allows us to assess how well our models generalize to specialized domains that differ significantly from the general Wikipedia text used during pre-training.

Table \ref{tab:ood_results} presents the F1-scores achieved on the MAPA dataset. The results reveal several interesting patterns regarding the adaptability of multilingual models to domain-specific tasks. Among the baseline models, XLM-RoBERTa-base shows the highest overall performance in the "all" configuration with an average F1-score of 73.04\%, followed by mBERT-base at 70.77\% and Distil-mBERT-base at 70.55\%. Interestingly, XLM-RoBERTa-large, despite its larger capacity, achieves a slightly lower average score of 69.51\% in the joint training scenario, suggesting that model size alone does not guarantee better domain adaptation.

The most notable finding is the significant improvement that all models, especially our UniBERT variants, experience when transitioning from individual language fine-tuning to multilingual joint  training. For instance, UniBERT-Small improves from an average F1-score of 46.78\% in the "each" setting to 64.63\% in the "all" setting, representing a relative improvement of 38.16\%. Similarly, UniBERT-XSmall and UniBERT-XXSmall show substantial gains, increasing from 38.18\% to 57.79\% (51.36\% relative improvement) and from 30.21\% to 47.04\% (55.71\% relative improvement), respectively.

This pronounced improvement in the legal domain suggests that multilingual joint  training and adversarial objectives are particularly beneficial for domain adaptation tasks. By learning language-invariant representations during pre-training, the models develop a more robust understanding of entity structures that transfers well across both languages and domains. With its specialized terminology and structured document formats, the legal domain benefits significantly from the cross-lingual representations learned through our training methodology.

\begin{table*}
    \centering

    \caption{Named entity recognition results on the MAPA dataset for the legal domain. We report the model's F1-score for each language (i.e., case "each") and for all languages at the same time (i.e., case "all"), evaluated on each language, together with their average F1-score.}

    \resizebox{\textwidth}{!}{
    \begin{tabular}{|l|c|c|c|c|c|c|c|c|c|c|c|c|c|c|c|c|c|c|c|c|c|c|c|}
         \toprule
         \textbf{Model} & \textbf{Train} & \textbf{bg} & \textbf{cs} & \textbf{da} & \textbf{de} & \textbf{el} & \textbf{en} & \textbf{es} & \textbf{et} & \textbf{fi} & \textbf{fr} & \textbf{ga} & \textbf{hu} & \textbf{it} & \textbf{lt} & \textbf{lv} & \textbf{mt} & \textbf{nl} & \textbf{pt} & \textbf{ro} & \textbf{sk} & \textbf{sv} & \textbf{Avg} \\
         \midrule
          \multirow{3}{*}{Distil-mBERT-base \cite{Sanh2019DistilBERTAD}} & each & 67.95 & 60.73 & 58.04 & 62.56 & 70.28 & 55.17 & 69.11 & 47.79 & 60.52 & 54.50 & 54.51 & 64.86 & 88.05 & 58.88 & 57.50 & 39.44 & 63.88 & 58.01 & 65.05 & 52.03 & 68.77 & 60.83 \\
         & all & 69.97 & 66.80 & 68.81 & 72.20 & 74.65 & 71.65 & 87.99 & 49.78 & 62.45 & 67.06 & 67.39 & 74.46 & 86.00 & 65.92 & 63.14 & 75.70 & 72.60 & 80.33 & 77.80 & 59.83 & 67.13 & 70.55 \\
         \midrule
         \multirow{2}{*}{mBERT \cite{devlin-etal-2019-bert}} & each & 73.50 & 67.96 & 68.52 & 67.86 & 70.89 & 60.17 & 77.11 & 54.26 & 66.55 & 67.69 & 58.85 & 74.54 & 78.18 & 66.79 & 62.42 & 59.39 & 69.57 & 75.44 & 65.09 & 61.54 & 69.51 & 67.42 \\
         & all & 73.85 & 77.22 & 64.26 & 68.45 & 67.16 & 76.77 & 91.82 & 56.15 & 62.98 & 60.35 & 59.26 & 76.89 & 87.97 & 66.13 & 62.96 & 80.21 & 75.06 & 73.35 & 74.59 & 66.51 & 64.33 & 70.77 \\
         \midrule
         \multirow{2}{*}{XLM-RoBERTa-base \cite{conneau2019cross}} & each & 66.25 & 60.40 & 63.94 & 57.05 & 50.87 & 52.66 & 49.06 & 46.46 & 64.47 & 48.87 & 53.33 & 56.51 & 89.20 & 52.67 & 49.25 & 34.82 & 61.47 & 56.17 & 59.35 & 50.85 & 72.61 & 56.96 \\
         & all & 66.53 & 68.55 & 76.34 & 77.03 & 78.30 & 70.52 & 91.39 & 64.58 & 70.03 & 61.51 & 68.76 & 72.28 & 86.08 & 73.68 & 68.71 & 71.72 & 71.88 & 81.34 & 76.89 & 61.98 & 75.80 & 73.04 \\
         \midrule
         \multirow{2}{*}{XLM-RoBERTa-large \cite{conneau2019cross}} & each & 62.67 & 62.48 & 71.31 & 60.03 & 62.33 & 72.67 & 76.92 & 50.87 & 68.72 & 64.05 & 56.66 & 74.70 & 77.25 & 62.45 & 44.55 & 57.51 & 69.00 & 47.14 & 70.11 & 46.92 & 77.86 & 63.62 \\
         & all & 66.27 & 64.65 & 69.96 & 73.33 & 70.49 & 72.39 & 88.59 & 50.50 & 61.94 & 58.70 & 62.96 & 73.44 & 90.03 & 71.72 & 67.54 & 77.53 & 67.38 & 70.91 & 75.28 & 60.43 & 65.72 & 69.51 \\
         \midrule
         \multirow{2}{*}{XLM-V-base \cite{liang2023xlm}} & each & 65.14 & 62.33 & 56.20 & 58.76 & 55.32 & 68.95 & 76.21 & 49.65 & 61.81 & 53.47 & 50.88 & 65.94 & 79.38 & 53.92 & 58.76 & 62.11 & 63.27 & 61.48 & 66.85 & 57.72 & 59.26 & 61.30 \\
     & all & 70.97 & 68.51 & 60.74 & 62.18 & 60.51 & 77.84 & 89.44 & 53.98 & 65.38 & 58.39 & 54.30 & 72.60 & 84.81 & 59.24 & 64.14 & 79.74 & 69.89 & 68.90 & 73.61 & 64.05 & 64.04 & 67.77 \\
     \midrule
     \multirow{2}{*}{\textit{UniBERT-small} (ours)} & each & 57.53 & 44.53 & 49.72 & 49.66 & 51.84 & 45.16 & 47.15 & 29.73 & 44.66 & 44.26 & 45.20 & 57.62 & 70.67 & 42.66 & 38.37 & 7.52 & 55.79 & 43.69 & 59.39 & 45.91 & 51.39 & 46.78 \\
     & all & 70.09 & 63.95 & 60.15 & 62.55 & 67.49 & 68.10 & 81.58 & 48.26 & 60.15 & 57.20 & 52.42 & 72.18 & 84.79 & 58.92 & 57.48 & 65.27 & 71.46 & 64.83 & 71.26 & 61.71 & 57.41 & 64.63 \\
     \midrule
     \multirow{2}{*}{\textit{UniBERT-xsmall} (ours)} & each & 50.83 & 38.16 & 39.47 & 39.65 & 42.08 & 47.38 & 50.03 & 22.36 & 38.58 & 20.30 & 23.01 & 45.98 & 73.47 & 41.35 & 35.44 & 0.00 & 56.04 & 41.16 & 51.76 & 2.25 & 42.54 & 38.18 \\
     & all & 69.83 & 63.38 & 51.93 & 51.34 & 62.80 & 58.86 & 71.25 & 37.48 & 56.93 & 53.10 & 45.21 & 59.46 & 73.09 & 53.39 & 49.74 & 57.63 & 65.22 & 54.28 & 63.33 & 63.18 & 51.17 & 57.79 \\
     \midrule
     \multirow{2}{*}{\textit{UniBERT-xxsmall} (ours)} & each & 39.85 & 34.56 & 36.27 & 32.14 & 37.93 & 14.82 & 15.64 & 17.93 & 32.45 & 14.26 & 19.58 & 38.29 & 66.47 & 30.18 & 27.35 & 10.42 & 46.32 & 31.03 & 41.58 & 12.17 & 35.24 & 30.21 \\
     & all & 53.95 & 51.07 & 48.03 & 45.89 & 47.34 & 53.10 & 65.71 & 23.94 & 44.07 & 46.08 & 38.75 & 48.88 & 69.35 & 34.06 & 33.94 & 14.94 & 61.93 & 42.97 & 60.25 & 54.31 & 49.35 & 47.04 \\
     \bottomrule
    \end{tabular}
    }
    \label{tab:ood_results}
\end{table*}

\subsection{Discussion}

\begin{figure*}
    \centering
    \includegraphics[width=0.8\linewidth]{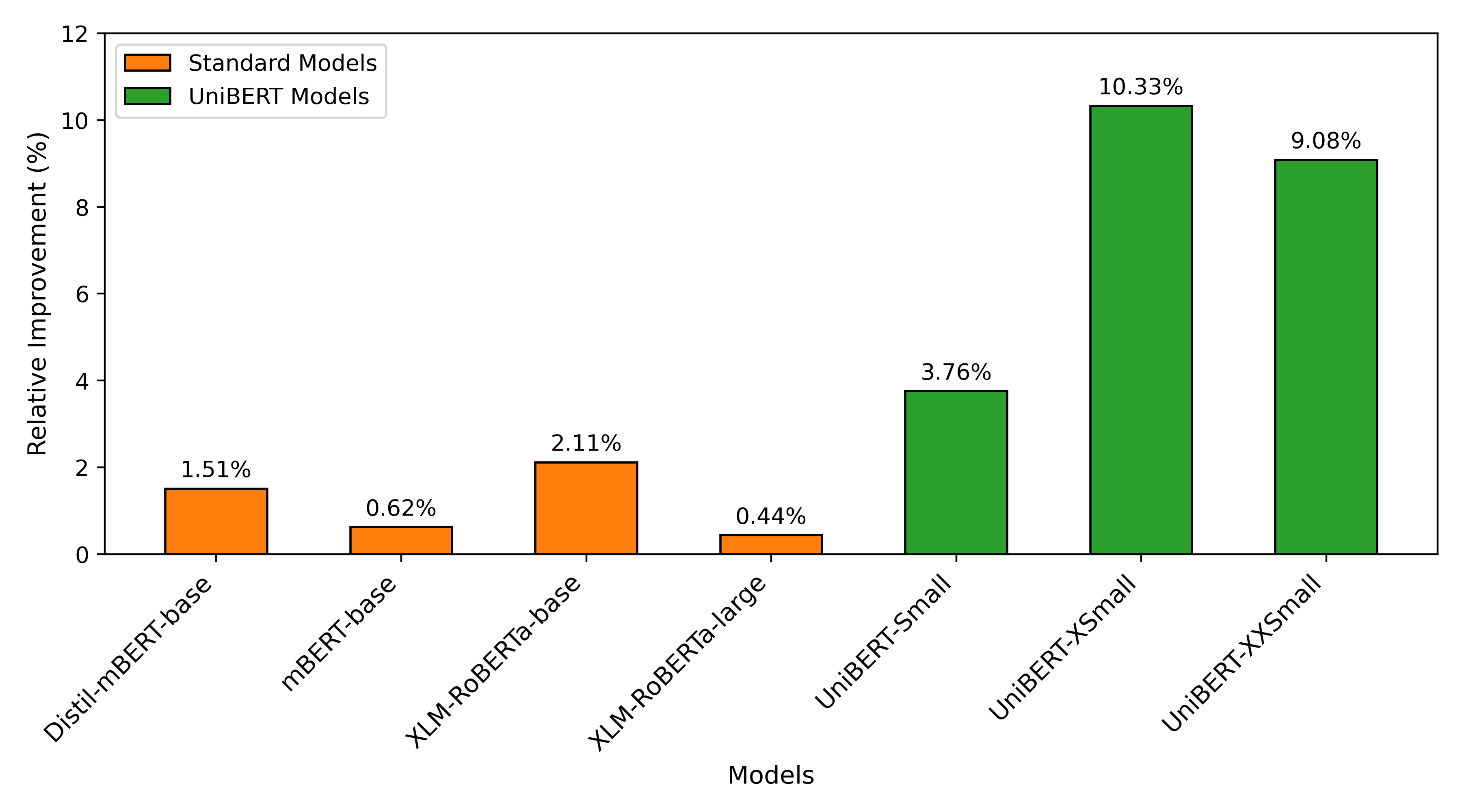}
    \caption{Relative improvements in UniBERT and other models when trained on individual languages (i.e., case "each") as opposed to all languages (i.e., case "all").}
    \label{fig:comparative_results}
\end{figure*}

The results indicate that the multilingual training strategy, in conjunction with the adversarial objective, has a pronounced impact on overall performance. Figure~\ref{fig:comparative_results} illustrates the relative improvements achieved by each model when transitioning from individual language fine-tuning (denoted as "each") to multilingual joint  fine-tuning (denoted as "all")\footnote{Since the score on the STS is between 0 and 1, we normalized it before calculating the mean relative improvement.}. mBERT-base, Distil-mBERT-base, and the XLM-RoBERTa versions exhibit a moderate average relative improvement of approximately 1.12\%: 1.51\% for Distil-mBERT-base, 0.62\% for mBERT-base, 2.11\% for XLM-RoBERTa-base, 0.44\% for XLM-RoBERTa-large, and 0.95\% for XLM-V-base. In contrast, UniBERT models indicate significantly greater gains, with an average relative improvement of 7.72\%: 3.76\% for UniBERT-small, 10.33\% for uniBERT-XSmall, and 9.08\% for UniBERT-XXSmall.

This significant difference suggests that the adversarial training component, which promotes the learning of language-agnostic representations, is particularly effective when combined with multilingual joint  training. For example, the UniBERT-XSmall model improves from 40.20\% in the "each" configuration to 44.35\% in the "all" configuration, and the UniBERT-XXSmall model similarly increases from 34.51\% to 37.64\%. Such improvements are notably higher than those observed in larger models, where the increase is relatively modest (e.g., XLM-RoBERTa-base and XLM-RoBERTa-large improve by less than 1.2 and 0.3 percentage points, respectively).

Performance variations between model sizes can be attributed to several key factors. Architecturally, UniBERT models with fewer parameters face capacity limitations when trained in individual languages, creating greater potential for improvement through cross-lingual knowledge sharing. Larger models like XLM-RoBERTa already effectively capture language-specific features during individual training, limiting additional gains from joint training. The adversarial objective functions differently between model sizes. In smaller architectures, it serves as an effective regularize, preventing overfitting to language-specific patterns while promoting efficient parameter usage. The Wikipedia training corpus provides consistent cross-lingual patterns that smaller models leverage more effectively with adversarial guidance. This effect is most pronounced in tasks requiring deeper semantic processing, such as NLI and QA, where UniBERT-XSmall demonstrates relative improvements of 12.7\% and 28.5\% respectively, suggesting adversarial training helps compact models prioritize cross-lingual semantic features over language-specific patterns.

Furthermore, a statistical t-test resulted in a t-statistic of 2.714 with a p-value of 0.0184 when comparing the relative improvements between the UniBERT models (12 entries, average improvement of 7.72\%) and the other models (16 entries, average improvement of 1.12\%). The observed differences are statistically significant because the p-value is lower than the significance threshold of $\alpha=0.05$. This analysis underscores that combining multilingual joint training with the adversarial objective enhances absolute performance and significantly benefits models with limited capacity by effectively leveraging cross-lingual representations.

The domain adaptation capabilities of UniBERT models further validate our approach. When evaluated in the legal domain using the MAPA dataset, UniBERT models demonstrate remarkable improvements when trained jointly across languages, with relative gains of 38.16\% for UniBERT-Small, 51.36\% for UniBERT-XSmall, and 55.71\% for UniBERT-XXSmall. This suggests that language-invariant representations learned through adversarial training facilitate effective cross-domain knowledge transfer, particularly benefiting specialized domains where annotated data may be scarce for specific languages.

\section{Limitations and Future Work}

Despite the promising results achieved by UniBERT, our approach faces certain limitations, primarily stemming from computational constraints that influenced our model design choices. To ensure feasibility with available resources, we opted for compact architectures with fewer parameters than the larger models used for comparison. This design choice, while enabling efficiency and scalability, inherently limits the model's capacity to learn and retain complex linguistic representations. Consequently, UniBERT exhibits lower absolute performance, particularly in tasks requiring deep contextual understanding, such as question answering and semantic textual similarity.

Our current mitigation strategy is primarily based on multilingual joint training and adversarial learning, which have proven effective in compensating for reduced capacity. These techniques promote the learning of language-agnostic representations that generalize well across languages, allowing UniBERT to achieve competitive results despite its smaller size. However, these approaches do not eliminate the inherent limitations of reduced parameterization. Larger models maintain an advantage in absolute performance due to their ability to capture more intricate linguistic structures and relationships.

Several additional technical strategies could further address these limitations without necessarily increasing the size of the model. One promising approach is to implement more sophisticated knowledge distillation techniques, such as layer-wise distillation with attention matching \cite{sajedi2023datadam}, which could enable better transfer of representational capabilities from larger teacher models. Parameter-efficient fine-tuning methods, such as adapter modules, LoRA (Low-Rank Adaptation) \cite{hulora}, or BitFit \cite{zaken2022bitfit}, represent viable strategies to enhance task-specific performance while maintaining a compact base model. Furthermore, implementing targeted bias mitigation strategies \cite{pozzi2025mitigating} would help ensure fairness across languages without expanding the model parameters, potentially reducing representational disparities between high- and low-resource languages while maintaining the compact nature of the model.

Future technical research could pursue two complementary directions: further model compression and performance optimization. Additional model compression techniques, such as quantization \cite{shao2024dq}, pruning \cite{corrHanMD15}, and specialized knowledge distillation variants for multilingual settings, could further reduce the model size while preserving cross-lingual capabilities. Specialized fine-tuning distillation protocols targeting specific linguistic phenomena across language families \cite{zhou2023universalner,he2025study} could substantially improve performance in tasks that require deeper semantic understanding. Another promising research direction would involve systematically evaluating compact multilingual models on edge devices across various hardware configurations, such as measurements of inference latency, memory usage, and power consumption in real-world deployment scenarios, particularly in multilingual mobile applications.

Beyond technical improvements, future work could investigate broader ethical and practical considerations of compact multilingual models. Researchers may develop robust bias mitigation strategies specifically designed for multilingual contexts, including methods to identify and remediate language-specific biases that may be amplified during model compression. Studies examining representation disparities between high- and low-resource languages in compressed models compared to their larger counterparts would provide valuable insights into fairness issues. Furthermore, expanding training datasets beyond Wikipedia to include more diverse text sources and genres across all 107 languages, with particular emphasis on increasing representation of low-resource languages, could address fundamental data limitations. Such an expansion would benefit from careful data curation processes to ensure quality and balance, potentially incorporating linguistic annotations to enhance the model's understanding of language-specific phenomena.

\section{Conclusion}

In this paper, we introduce UniBERT, a multilingual language model that employs a novel training framework consisting of masked language modeling, adversarial training, and knowledge distillation. Our extensive evaluations of four NLP tasks (i.e., named entity recognition, natural language inference, question answering, and semantic textual similarity) demonstrate that the proposed approach achieves competitive performance relative to established baselines and benefits significantly from multilingual joint training. The UniBERT variants, for example, obtained an average relative improvement of 7.72\% between tasks. However, the larger baseline models showed an improvement of 1.12\%.

Integrating the adversarial objective has proven particularly effective in fostering language-invariant representations, enhancing cross-lingual generalization, and producing statistically significant gains (i.e., p-value = 0.0184) over models trained in individual languages when trained jointly on multilingual data. Furthermore, our out-of-domain evaluation of legal documents demonstrates that these language-invariant representations facilitate cross-domain generalization, with UniBERT models showing up to 55.71\% relative improvement when trained jointly across languages.

The promising results of UniBERT underscore the potential of our methodology in addressing the challenges associated with multilingual processing in resource-constrained settings. Although our model balances efficiency and performance, there remains ample scope for future work, including further refinement of adversarial strategies, exploration of alternative distillation techniques, and extension to additional languages and domains. Overall, the contributions of this work provide a robust foundation for future research in efficient multilingual modeling and adaptable and scalable NLP systems.

\section*{Acknowledgements}
This work was supported by the National University of Science and Technology POLITEHNICA Bucharest through the PubArt program. DCC is funded by the National Program for Research of the National Association of Technical Universities (GNAC ARUT 2023).

\bibliographystyle{unsrt}  
\bibliography{references}  

\end{document}